\documentclass[twocolumn]{article} % DO NOT CHANGE THIS
\usepackage[margin=0.8in]{geometry}
\usepackage{times}
\usepackage{helvet}
\usepackage{courier}
\usepackage[hyphens]{url}
\usepackage{hyperref}
\usepackage{graphicx}
\usepackage{caption}
\usepackage{algorithm}
\usepackage{algpseudocode}

\usepackage{amsmath} 
\usepackage{amssymb}

\usepackage{xcolor}

\usepackage{comment}
\usepackage{subcaption}

\newcommand{\fnoisysphere}{\tilde{f}_{\text{sphere}}}
\newcommand{\fnoisyelli}{\tilde{f}_{\text{elli}}}

\newcommand{\step}[1]{\texttt{step(\ensuremath{#1})}}

\title{Cumulative Learning Rate Adaptation: Revisiting Path-Based Schedules for SGD and Adam}
\author{
    Asma Atamna\textsuperscript{1},
    Tom Maus\textsuperscript{1},
    Fabian Kievelitz\textsuperscript{1},
    Tobias Glasmachers\textsuperscript{1} \\
    \textsuperscript{1}Ruhr University Bochum, Germany \\
    \texttt{\{firstname.lastname\}@rub.de}
}

\date{}

\begin{document}
\maketitle

%-----------------------------------%
\begin{abstract}
%-----------------------------------%

The learning rate is a crucial hyperparameter in deep learning, with its ideal value depending on the problem and potentially changing during training. In this paper, we investigate the practical utility of adaptive learning rate mechanisms that adjust step sizes dynamically in response to the loss landscape. We revisit a cumulative path-based adaptation scheme proposed in 2017, which adjusts the learning rate based on the discrepancy between the observed path length, computed as a time-discounted sum of normalized gradient steps, and the expected length of a random walk. While the original approach offers a compelling intuition, we show that its adaptation mechanism for Adam is conceptually inconsistent due to the optimizer’s internal preconditioning. We propose a corrected variant that better reflects Adam’s update dynamics. To assess the practical value of online learning rate adaptation, we benchmark SGD and Adam, with and without cumulative adaptation, and compare them to a recent alternative method. Our results aim to clarify when and why such adaptive strategies offer practical benefits.
\end{abstract}

%-----------------------------------%
\section{Introduction}
\label{sec:introduction}
%-----------------------------------%

Gradient-based optimization underpins modern machine learning, powering progress in domains ranging from image recognition to reinforcement learning~\cite{Goodfellow2016, Bottou2018, sutton2018}. Among the available methods, Stochastic Gradient Descent (SGD) remains a cornerstone due to its simplicity and scalability~\cite{Bottou2018}. However, its performance is highly sensitive to the choice of hyperparameters, particularly the learning rate. An inappropriate learning rate can slow convergence, cause divergence, or optimize traps in poor regions of the loss landscape~\cite{Smith2018}.

Adaptive optimizers such as AdaGrad~\cite{Duchi2011}, RMSProp~\cite{Hinton2012}, and Adam~\cite{Kingma2015} aim to mitigate these issues by scaling updates based on past gradient information. Adam, in particular, has become a default choice due to its strong empirical performance and relatively low sensitivity to gradient scale and initialization. Like other adaptive methods, it relies on a global learning rate whose choice remains critical~\cite{Kingma2015, Goodfellow2016, Smith2017}. In practice, this often necessitates heuristic schedules or exhaustive tuning.

Despite decades of research, learning rate selection remains one of the most impactful and under-specified aspects of training deep models~\cite{wu2023}. Manual tuning or fixed schedules are widely used, but these approaches are brittle: they are expensive to tune, often problem-specific, and can reduce reproducibility across architectures and datasets~\cite{Goyal2017, Smith2017}.

In this paper, we explore an alternative approach through Cumulative Learning Rate Adaptation (CLARA), a lightweight mechanism that adjusts the global learning rate on the fly by analyzing the optimizer’s trajectory. Rather than relying on instantaneous gradient magnitudes or predefined schedules, CLARA leverages the cumulative directionality of recent updates to infer whether current steps are consistently aligned or conflicting, guiding the learning rate accordingly.

Our contribution builds on a previous path length–based adaptation method for SGD~\cite{SchoenauerSebag2017}, extending it conceptually and algorithmically to accommodate adaptive optimizers like Adam. We correct a key inconsistency in how path lengths are interpreted under preconditioning and show how to adapt the reference comparison to Adam’s geometry. The result is a general mechanism that improves robustness without introducing significant computational cost.

We evaluate CLARA on synthetic optimization problems and supervised learning tasks spanning classical tabular datasets and high-dimensional image classification. Our empirical study addresses the following questions:

\begin{enumerate}
    \item Can CLARA improve robustness to learning rate initialization, particularly in settings where manual tuning is difficult or unreliable?
    \item How does CLARA interact with Adam’s internal adaptation? Can it still add value despite preconditioning?
    \item How does CLARA compare to other recent adaptation methods, such as D-Adaptation, in terms of stability and convergence?
\end{enumerate}

Overall, our results show that CLARA is most impactful when applied to SGD, where it recovers from poor initializations and mimics the benefits of hand-tuned schedules. When applied to Adam, its benefits are more context-dependent but still visible in low-signal or poorly initialized regimes. CLARA provides a simple and effective tool for online learning rate control, contributing toward more resilient and hands-free training strategies.

The remainder of this paper is organized as follows: Section~\ref{sec:related-work} reviews related work, Section~\ref{sec:algorithm} describes CLARA, Section~\ref{sec:experimental-eval} presents our experimental setup, Section~\ref{sec:results} discusses the results, and Section~\ref{sec:discussion} concludes the paper.

%-----------------------------------%
\section{Related Work}
\label{sec:related-work}
%-----------------------------------%

Recent work has increasingly focused on reducing the burden of learning rate tuning in gradient-based optimization. Here, we review classical adaptive methods, recent advances in online learning rate control and approaches that leverage information from optimizer trajectories. We also situate our contribution within this growing body of research.

\subsection{Adaptive Gradient Methods and Learning Rates}
Classical adaptive methods such as AdaGrad~\cite{Duchi2011}, RMSProp~\cite{Hinton2012} and Adam~\cite{Kingma2015} adjust learning rates based on gradient history, addressing challenges like sparse gradients and non-uniform curvature. These optimizers are widely used due to their empirical robustness, particularly Adam, which is effective across a range of deep learning tasks. However, they all require a global learning rate, whose manual tuning remains critical and costly~\cite{Smith2017}.

The need to automate learning rate selection has sparked a recent wave of research, marking a shift away from traditional offline tuning approaches toward more principled, online adaptation.
D-Adaptation~\cite{defazio2023} introduces a hyperparameter-free mechanism for learning rate adaptation that provably achieves optimal convergence rates for convex Lipschitz objectives. The key idea is to iteratively estimate a lower bound on the distance to the optimum and use this to update the learning rate without requiring line searches or additional gradient evaluations. A recent follow-up, Prodigy~\cite{mishchenko2024prodigy}, improves upon D-Adaptation by increasing the learning rate faster, making it preferable when D-Adaptation tends to underestimate the learning rate.
GALA~\cite{jiang2025gala} formulates learning rate adaptation as a one-dimensional online learning problem, updating step sizes based on gradient alignment and curvature estimates. By monitoring the agreement between the current gradient and an approximation of the average gradient along the recent optimization path, GALA increases the learning rate when progress is steady and decreases it when gradients become misaligned.
In~\cite{faraj2025cumulative}, the authors investigate the relationship between dataset size and optimal learning rates, revealing that the ideal learning rate is inversely proportional to the total number of data exposures (i.e., dataset size times number of epochs). This leads to the formulation of a cumulative learning constant, which serves as a guiding principle for designing and scaling learning rate schedules. Rather than adapting learning rates online, this work offers a systematic way to set global learning rates ahead of time and transfer them across training regimes of different scales.

\subsection{Trajectory-Based Adaptation}
In evolutionary optimization, step-size control via cumulative path statistics is a key ingredient of the state-of-the-art black-box optimizer CMA-ES~\cite{hansen2001}. SALERA transfers this idea to SGD by comparing the length of a running sum of normalized gradients to that of a random walk~\cite{SchoenauerSebag2017}. While promising, SALERA was limited to unconditioned SGD and relied on an additional change point detection recovery mechanism. 

\subsection{Our Contribution}
Our method, CLARA, builds on and generalizes the cumulative path-based adaptation principle, proposing a simpler, continuously adaptive mechanism applicable to preconditioned optimizers like Adam. We situate CLARA within the active research area on adaptive step-size control, aiming to clarify when and why cumulative adaptation improves training. In contrast to prior work, CLARA introduces a unified, trajectory-aware rule for learning rate adjustment that works for both SGD and Adam. Our method is lightweight, first-order, and does not require auxiliary gradient evaluations or loss modeling, making it a practical tool for improving robustness in modern machine learning training pipelines.

\section{Algorithm}
\label{sec:algorithm}
%-----------------------------------%

Our proposed approach, Cumulative Learning Rate Adaptation (CLARA), modifies the Safe Agnostic Learning Rate Adaptation (SALERA) method~\cite{SchoenauerSebag2017}, which draws inspiration from the Covariance Matrix Adaptation Evolution Strategy (CMA-ES)~\cite{hansen2001}, a well-established derivative-free optimizer. 

\begin{figure}
    \centering
    \includegraphics[width=1\linewidth]{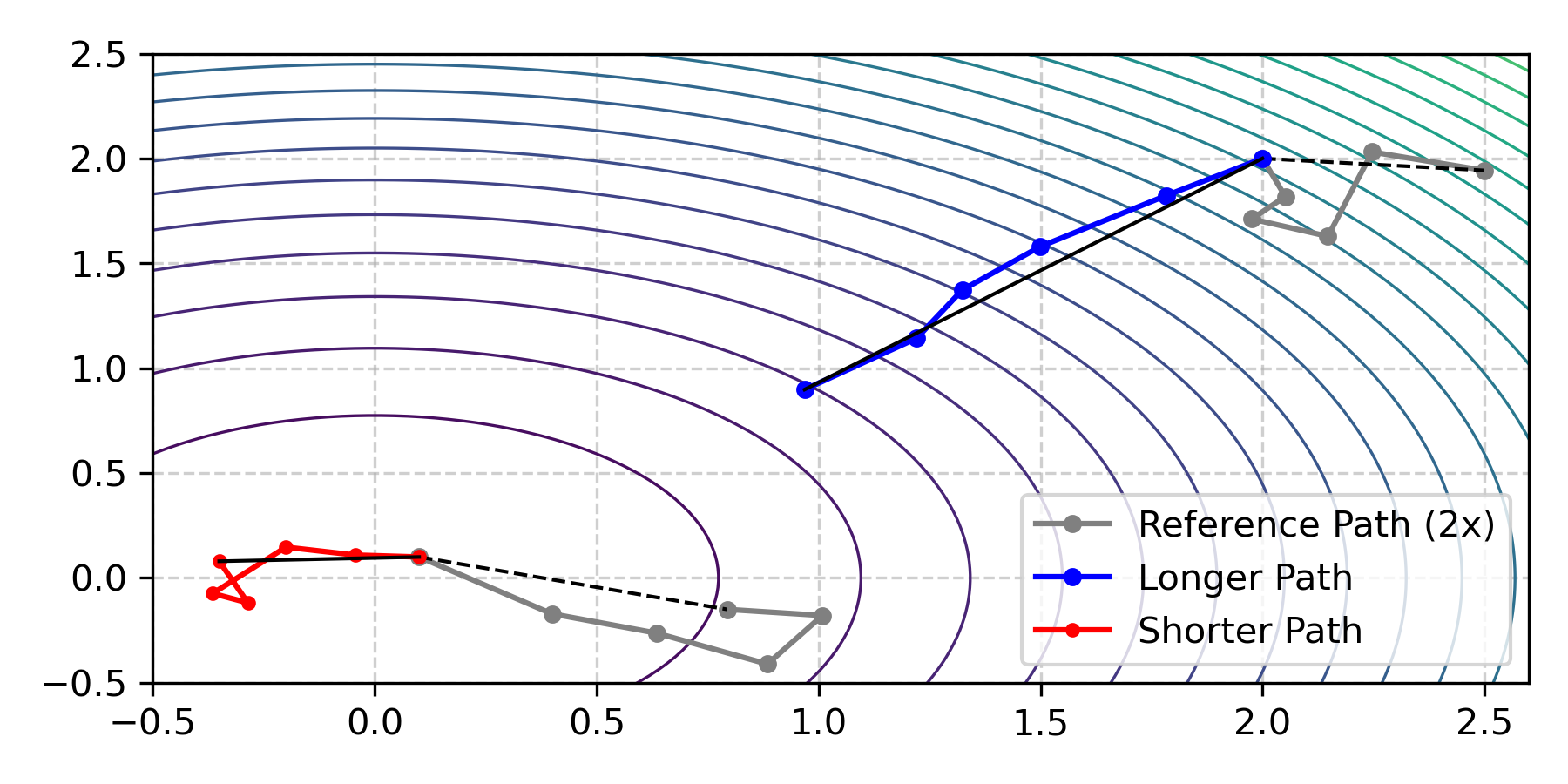}
    \caption{Cumulative paths in blue (top right) and red (bottom left) compared to random reference paths in gray. Path length is measured from the first to the last node, shown as solid black lines for the optimizer's paths and dashed black lines for the random references.}
    \label{fig:cumulative-path-visualization}
\end{figure}

CLARA adjusts the learning rate on the fly based on the optimizer’s step history in the parameter space, referred to as the \textit{cumulative path}. The core idea is to relate the length of this path to that of a reference path generated by taking consecutive random steps: a longer-than-expected path suggests consistent directions and slow progress, calling for an increased learning rate; a shorter path suggests conflicting directions and potential overshooting, warranting a decrease. Figure~\ref{fig:cumulative-path-visualization} provides visual intuition.

While SALERA computes the cumulative path as a sum of normalized gradients, which is appropriate for SGD, it does not account for the internal preconditioning used by Adam. CLARA addresses this limitation by computing the cumulative path from Adam’s normalized preconditioned steps and applying the same preconditioning to the reference random steps.

Algorithm~\ref{alg:generalized_clara} summarizes CLARA applied to a general stochastic gradient-based optimizer. The key steps of the procedure are described in detail below.

\subsection{Step Update}
At each iteration, the algorithm computes the gradient $g_t = \nabla_x f(x_t)$ and uses a base optimizer (e.g., SGD or Adam) to compute the step direction $s_t = \step{g_t}$. The function \texttt{step} is defined as the identity for SGD, and follows the standard update based on first- and second-order moment estimates for Adam:
\begin{align*}
m_{t + 1} &= \beta_1 m_t + (1 - \beta_1) g_t , \\
\ v_{t + 1} &= \beta_2 v_t + (1 - \beta_2) g_t^2 , \\
\hat{m}_{t + 1} &= m_{t + 1} / (1 - \beta_1^{t + 1}) , \\
\hat{v}_t &= v_t / (1 - \beta_2^{t + 1}) , \\
s_t &= \hat{m}_{t + 1} / (\sqrt{\hat{v}_{t + 1}} + \varepsilon) .
\end{align*}
This step is optionally normalized depending on a boolean flag \texttt{unit\_step}. If \texttt{unit\_step} is true, the update $x_{t + 1} = x_t - \eta_t \hat{s_t}$ uses a unit-norm direction $\hat{s}_t = s_t / \|s_t\|$; otherwise, the full-length step $s_t$ is used.
Using unit-norm steps decouples the direction provided by the optimizer from the step magnitude, allowing CLARA to control only the global learning rate. This can improve stability and interpretability by ensuring that all step size adaptation is governed explicitly by CLARA.

\subsection{Cumulative Path Tracking}
The cumulative path $p_t$ is an exponential moving average of the normalized steps:
$$p_{t + 1} = (1 - c) p_t + c \hat{s}_t ,$$
where $c \in (0, 1]$ is a smoothing parameter. The cumulative path captures the directional consistency of recent updates: a long cumulative path implies aligned steps, while a short path indicates conflicting directions. Unlike SALERA, which only uses normalized gradient directions, CLARA tracks the normalized directions of the actual update steps, reflecting the transformation applied by the underlying optimizer (e.g., Adam).

\subsection{Learning Rate Adaptation}
To determine whether the cumulative path length $\|p_{t + 1}\|^2$ indicates unusually consistent or conflicting updates, CLARA compares it to the expected squared norm of a reference path built from random directions. This reference serves as a non-redundant baseline: in high-dimensional spaces, independently drawn directions are nearly orthogonal, causing their cumulative sum to grow steadily. Such behavior reflects an idealized scenario of non-conflicting progress against which CLARA gauges alignment.
% Figure~\ref{fig:cumulative-path-visualization} provides visual intuition: a cumulative path longer than the reference suggests consistently aligned steps and triggers a learning rate increase; a shorter path implies conflicting directions and leads to a decrease.

The definition of the reference path depends on the base optimizer. For SGD, it consists of uniformly drawn unit vectors, and its expected norm can be computed analytically~\cite{SchoenauerSebag2017}. For Adam, which applies direction-changing preconditioning based on first and second moment estimates, the reference path is constructed by summing normalized Gaussian vectors that are transformed using the same preconditioning scheme.
Since no closed-form expression is available for this setup, the expected squared norm, denoted $\mathbb{E}(\|r_{t + 1}\|^2)$, is estimated empirically using a Monte Carlo simulation as described in Algorithm~\ref{alg:reference-path-norm-estimation-adam}.
The learning rate is updated as follows:
$$\eta_{t+1} = \eta_t \cdot \exp\left(d \left( \frac{\|p_{t+1}\|^2}{\mathbb{E}(\|r_{t+1}\|^2)} - 1 \right) \right) ,$$
where $0 < d \leq 1$ is a damping parameter that controls the responsiveness of the adaptation. This rule increases the learning rate when the observed path is longer than expected, and decreases it when the path is shorter.
\begin{algorithm}\caption{Generalized Optimizer with CLARA}
\label{alg:generalized_clara}
\begin{algorithmic}[1]
\Require Learning rate $\eta_0 > 0$, smoothing parameter $0 < c \leq 1$, damping parameter $0 < d \leq 1$, initial point $x_0$, function $f$, boolean \texttt{unit\_step}, estimate of $\mathbb{E}(\|r_{t+1}\|^2)$
\State Initialize timestep $t = 0$, $p_0 = 0$
\While{not converged}
    \State $g_t = \nabla_x f(x_t)$ \hfill \Comment{Compute gradient}
    \State $s_t = \step{g_t}$ \hfill \Comment{Compute update direction}
    \State $\hat{s}_t = s_t / \|s_t\|$ \hfill \Comment{Normalize direction}
    \If{\texttt{unit\_step}}
        \State $x_{t + 1} = x_t - \eta_t \hat{s}_t$ \hfill \Comment{Take unit-norm step}
    \Else
        \State $x_{t + 1} = x_t - \eta_t s_t$ \hfill \Comment{Take full-length step}
    \EndIf
    \State $p_{t + 1} = (1 - c) p_t + c \hat{s}_t$ \hfill \Comment{Compute cumulative path}
    \State $\eta_{t+1} = \eta_t \cdot \exp\left(d \left( \frac{\|p_{t+1}\|^2}{\mathbb{E}(\|r_{t+1}\|^2)} - 1 \right) \right)$ \hfill \Comment{Update learning rate}
    \State $t = t + 1$ \hfill \Comment{Update timestep}
\EndWhile
\State \textbf{Return} $x_t$
\end{algorithmic}
\end{algorithm}
\begin{algorithm}
\caption{Monte Carlo $\mathbb{E}(\|r_{t + 1}\|^2)$ Estimation for Adam}
\label{alg:reference-path-norm-estimation-adam}
\begin{algorithmic}[1]
\Require $\beta_1 = 0.9$, $\beta_2 = 0.999$, $\varepsilon = 10^{-8}$, $0 < c \leq 1$
\State Initialize $\mu = 0$ \hfill \Comment{Expectation estimate}
\For{trial $= 1$ to $N$}
    \State Initialize $m = 0$, $v = 0$, $r = 0$
    \For{$t = 1$ to $T$}
        \State $\xi_t \sim \mathcal{N}(0, I)$ \hfill \Comment{Sample Gaussian vector}
        \State $m = \beta_1 m + (1 - \beta_1) \xi_t$
        \State $v = \beta_2 v + (1 - \beta_2) \xi_t^2$
        \State $\hat{m} = m / (1 - \beta_1^t)$
        \State $\hat{v} = v / (1 - \beta_2^t)$
        \State $s = \hat{m} / (\sqrt{\hat{v}} + \varepsilon)$ \hfill \Comment{Simulated Adam step}
        \State $\hat{s} = s / \|s\|$ \hfill \Comment{Normalize direction}
        \State $r = (1 - c) r + c \hat{s}$ \hfill \Comment{Exponential moving average}
    \EndFor
    \State $\mu = \mu + \|r\|^2$
\EndFor
\State \textbf{Return} $\mu / N$
\end{algorithmic}
\end{algorithm}

%-----------------------------------%
\section{Experimental Evaluation}
\label{sec:experimental-eval}
%-----------------------------------%

The goal of our experiments is to validate the algorithmic ideas behind CLARA by comparing its performance to widely used optimization methods, specifically SGD and Adam, as well as the more recent D-Adaptation approach. We evaluate on both synthetic benchmarks and standard supervised learning tasks. While no method can consistently outperform a well-tuned fixed learning rate, especially when the ideal schedule is static, we investigate whether CLARA improves robustness and convergence in settings where tuning is difficult or the optimal rate evolves.

\subsection{Synthetic Benchmarks}
To gain initial insight into the behavior of the proposed learning rate adaptation strategy, we evaluate it on two widely used test functions from numerical optimization: the sphere function and the ill-conditioned ellipsoid function, both defined over $\mathbb{R}^n$. We focus on noisy variants of these functions, denoted $\fnoisysphere$ and $\fnoisyelli$, where Gaussian noise is added to the input before evaluation. This introduces stochasticity that mimics the kind of gradient noise often encountered in practical training scenarios. The mathematical definitions follow standard forms~\cite{hansen2021coco}:
\begin{align*}
    \fnoisysphere(x) &= \sum_{i = 1}^n (x_i + \xi_i)^2, \quad \xi_i \sim \mathcal{N}(0, \sigma^2) , \\
    \fnoisyelli(x) &= \sum_{i = 1}^n \alpha^{\frac{i - 1}{n - 1}} (x_i + \xi_i)^2, \quad \xi_i \sim \mathcal{N}(0, \sigma^2) ,
\end{align*}
with condition number $\alpha = 10^3$ and noise standard deviation $\sigma = 0.1$.

These benchmarks provide a controlled testbed for studying convergence behavior, sensitivity to learning rate initialization, and robustness to noise. To qualitatively assess the dynamics, we visualize optimizer trajectories using contour plots of the loss surfaces in two dimensions.

\subsection{Supervised Learning Benchmarks}
We conduct systematic experiments on a diverse collection of supervised learning tasks. These experiments encompass both classical tabular datasets and standard image classification benchmarks commonly used in the evaluation of optimization algorithms.

The tabular datasets include the UCI Breast Cancer, Iris, Wine, and Digits datasets, which are well-established in the machine learning community for benchmarking classification algorithms under constrained data regimes~\cite{Dua2017}. For high-dimensional and structurally richer inputs, we additionally consider MNIST and Fashion-MNIST~\cite{LeCun1998, Xiao2017}, as well as CIFAR-10 and CIFAR-100~\cite{Krizhevsky2009}. These datasets vary significantly in sample complexity, class imbalance, and input dimensionality, thus providing a broad empirical testbed for optimizer comparison.

Each dataset is associated with a suitable model architecture: logistic regression and shallow neural networks are used for the tabular datasets, while convolutional neural networks are employed for image classification tasks. Training is carried out using mini-batch stochastic optimization with a batch size of 128 for a total of 100 epochs. Performance is measured in terms of classification accuracy on held-out test data and averaged over 5 random seeds ranging from 0 to 4 to account for variance due to stochasticity. To assess sensitivity to learning rate initialization, each experiment is repeated across a logarithmically spaced grid of values $\eta_0 \in \{10^{-6}, 10^{-5}, \ldots, 1\}$. This setup allows us to analyze the stability and performance envelope of each optimizer.

The set of optimizers includes standard SGD and Adam~\cite{Kingma2015}, as well as D-Adaptation~\cite{defazio2023} and our proposed CLARA-enhanced variants of SGD and Adam. We additionally include unit step versions of the latter to study the influence of conservative adaptation. All optimizers are implemented via a centralized factory interface that ensures consistency and comparability across experiments.

\subsection{Implementation Details}
All experiments are implemented in \texttt{PyTorch} (v2.1, CUDA~11.8).  
Optimizers, data pipelines, training loops, and evaluation logic are separated into dedicated modules, enabling plug-and-play replacement of optimizers and parallel hyperparameter sweeps via Python’s \texttt{multiprocessing} backend. Unless explicitly stated otherwise, the default hyperparameter values provided by the corresponding package or \texttt{PyTorch} implementation were used for SGD, Adam, and D-Adaptation.
The full codebase used in our experiments is publicly available at \href{https://github.com/asmaatamna/cumulative-learning-rate-adaptation}{\texttt{GitHub}}.

\paragraph{CLARA hyperparameters}
Each CLARA variant uses a fixed smoothing factor of $c = 0.2$ and a damping factor $d$, which controls the responsiveness of the learning rate adaptation. For the comparative performance plots where we evaluate all tested algorithms, we select the best-performing damping value for each dataset and initial learning rate from a sweep over $d \in \{10^{-5}, 10^{-4}, \ldots, 10^{-1}\}$. By default, the damping is set to $10^{-3}$, a value that performs reasonably well across tasks and often yields results comparable to (and occasionally better than) those of the corresponding base optimizers without CLARA. For CLARA with Adam, the expected squared norm of the reference path is estimated as described in Algorithm~\ref{alg:reference-path-norm-estimation-adam} using $N = 1000$ and $T = 1000$, with the estimate depending on the dimensionality of the parameter space. For CLARA with SGD, we use the theoretical expression derived in~\cite{SchoenauerSebag2017} for SALERA.

\paragraph{Unit-step mode}  
A boolean flag activates a unit-step variant that normalizes the raw update to unit length before parameter changes, so that CLARA only modulates the global learning rate.  
Both default and unit-step (``us'') versions are benchmarked in our experiments.  

\paragraph{Parameter group handling}
While Algorithm~\ref{alg:generalized_clara} describes CLARA at the level of a single parameter vector, our implementation follows the parameter group structure used in \texttt{PyTorch} optimizers. The step direction and cumulative path are computed separately for each parameter group. For learning rate adaptation, the group-wise cumulative paths are flattened and concatenated before computing their squared norm. To ensure proper scaling, this norm is normalized by the expected reference value multiplied by the number of parameter groups.

%-----------------------------------%
\section{Results}
\label{sec:results}
%-----------------------------------%

We evaluate CLARA variants on both synthetic and supervised benchmarks, comparing them to SGD, Adam, and D-Adaptation. Results focus on robustness to learning rate initialization, convergence, and adaptation dynamics.

\subsection{Synthetic Benchmarks} \label{subsec:res-synthetic-benchmarks}
\begin{figure}[htbp]
    \centering
    \includegraphics[width=0.49\linewidth, trim=8pt 10pt 20pt 10pt, clip]{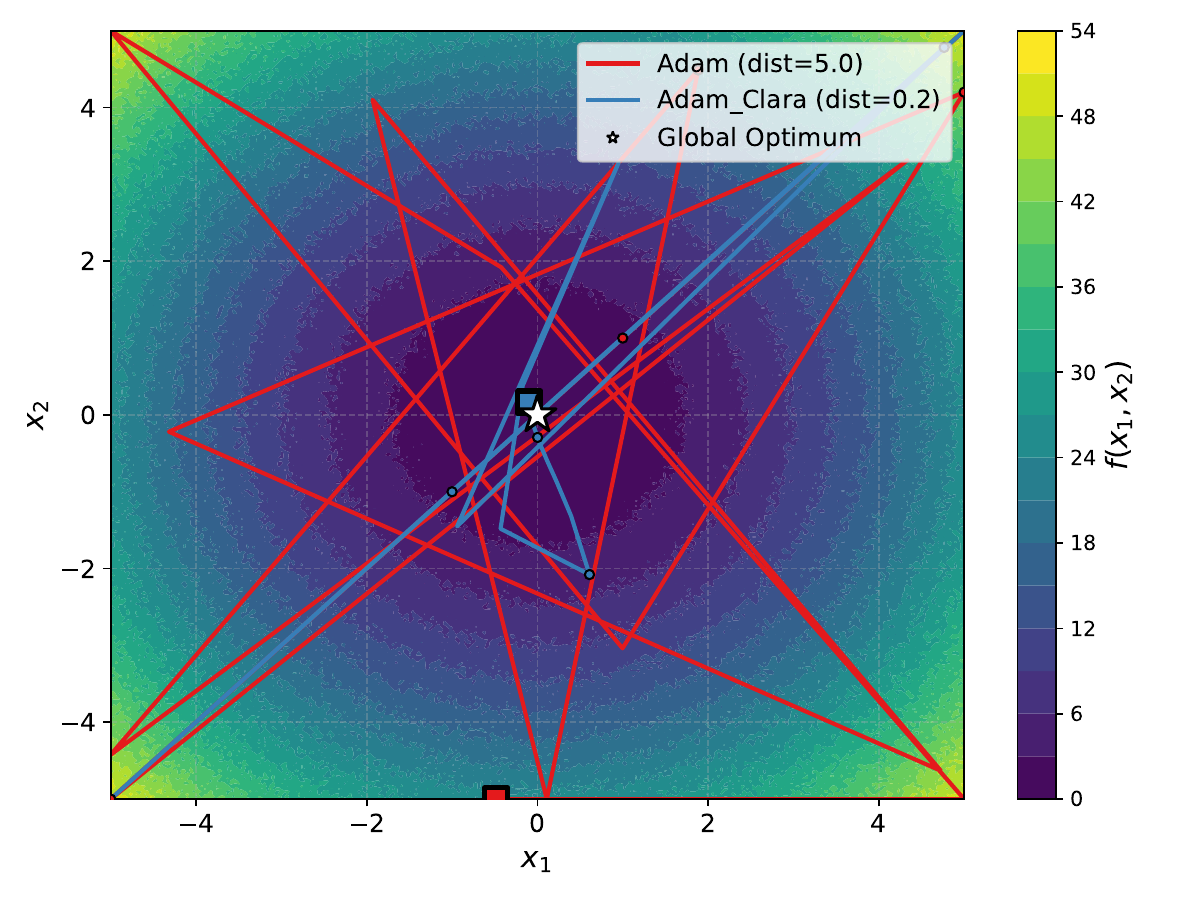}
    \includegraphics[width=0.49\linewidth, trim=8pt 10pt 20pt 10pt, clip]{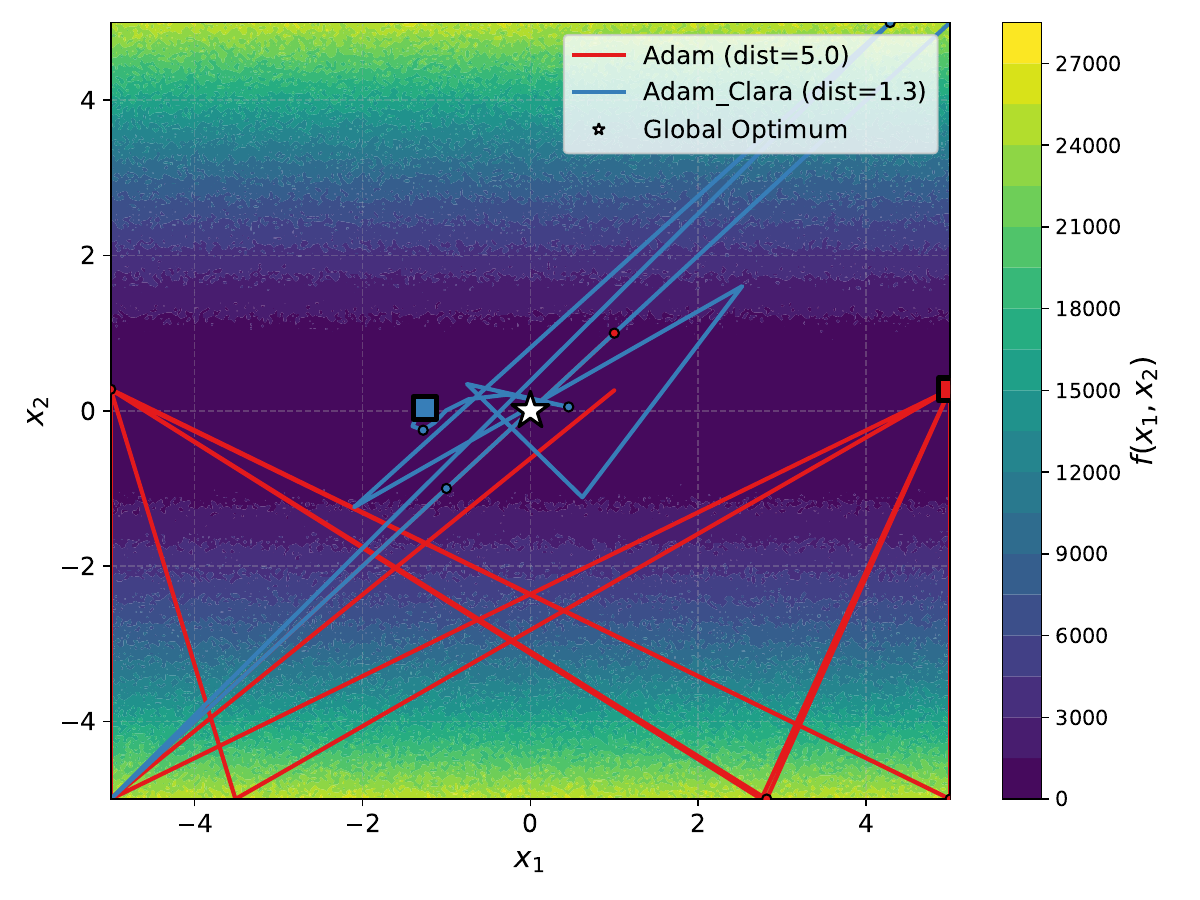}
    \caption{Trajectories of Adam and Adam\_CLARA on $\fnoisysphere$ (left) and $\fnoisyelli$ (right) with $\eta_0 = 10^2$. Each plot shows contour lines of the objective and consecutive optimization steps. A square marker indicates the final point reached after the optimization budget is exhausted. The legend also reports the final distance to the optimum for each optimizer. While Adam diverges due to the large learning rate, CLARA gradually reduces it and guides the optimizer toward the optimum despite the noise.}
    \label{fig:noisy_sphere_elli_trajectory}
\end{figure}

Figure~\ref{fig:noisy_sphere_elli_trajectory} depicts the optimization trajectories of Adam and Adam\_CLARA on the two–dimensional noisy sphere ($\fnoisysphere$) and noisy ellipsoid ($\fnoisyelli$) functions for a relatively large initial learning rate value $\eta_0 = 10^2$.  
Both methods eventually reach the vicinity of the global optimum, yet the cumulative-adapted variant travels a markedly straighter path and arrives in fewer iterations.  
Across both synthetic benchmarks, CLARA significantly reduces the final distance to the optimum compared to its non-adaptive counterpart. On $\fnoisysphere$, CLARA converges within a radius of 0.2, whereas Adam overshoots and stabilizes around a distance of 5. On $\fnoisyelli$, CLARA reaches within 1.3 units of the optimum, while Adam remains at a much larger distance of 5. These improvements suggest that trajectory-based learning rate control in CLARA helps mitigate overshooting caused by high curvature or large initial learning rates. This tendency was consistent across multiple independent runs.

\subsection{Supervised Learning Benchmarks}

We organize the results of the supervised benchmarks in two parts: a comparative view across benchmarked algorithms and a closer look at the learning rate adaptation dynamics induced by CLARA.

\begin{figure}[t]
    \centering
    \includegraphics[width=0.49\linewidth]{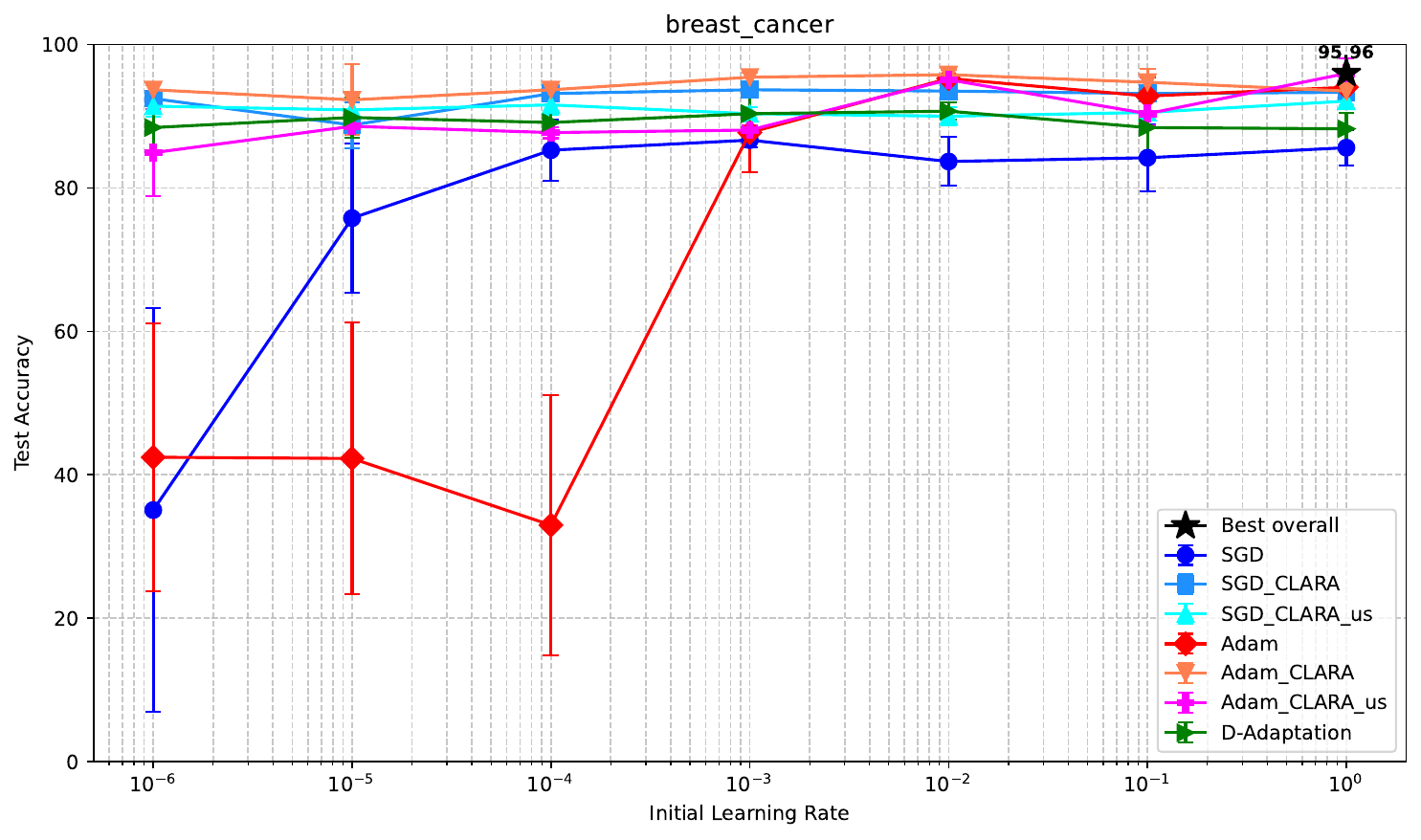}
    \includegraphics[width=0.49\linewidth]{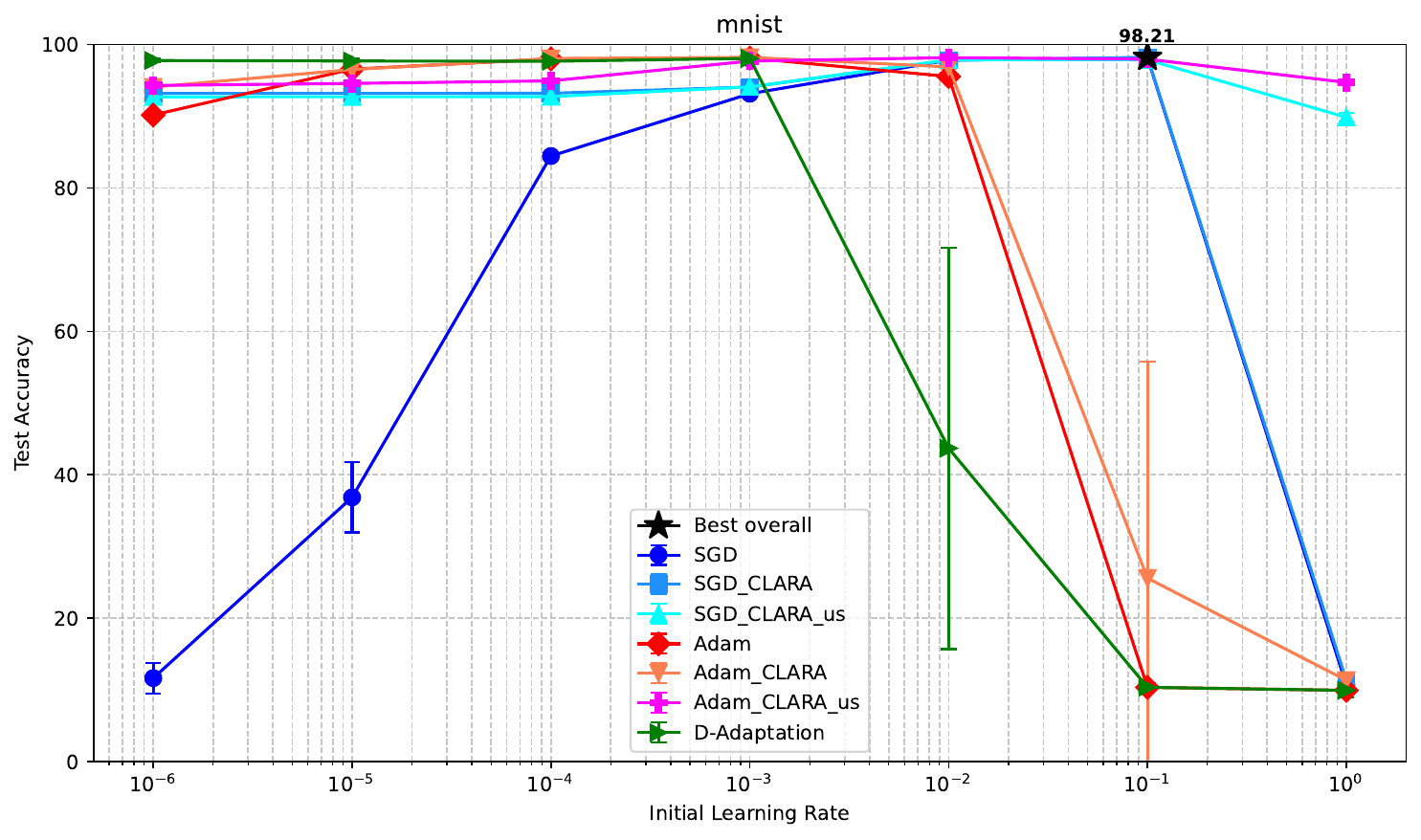} \\
    \includegraphics[width=0.49\linewidth]{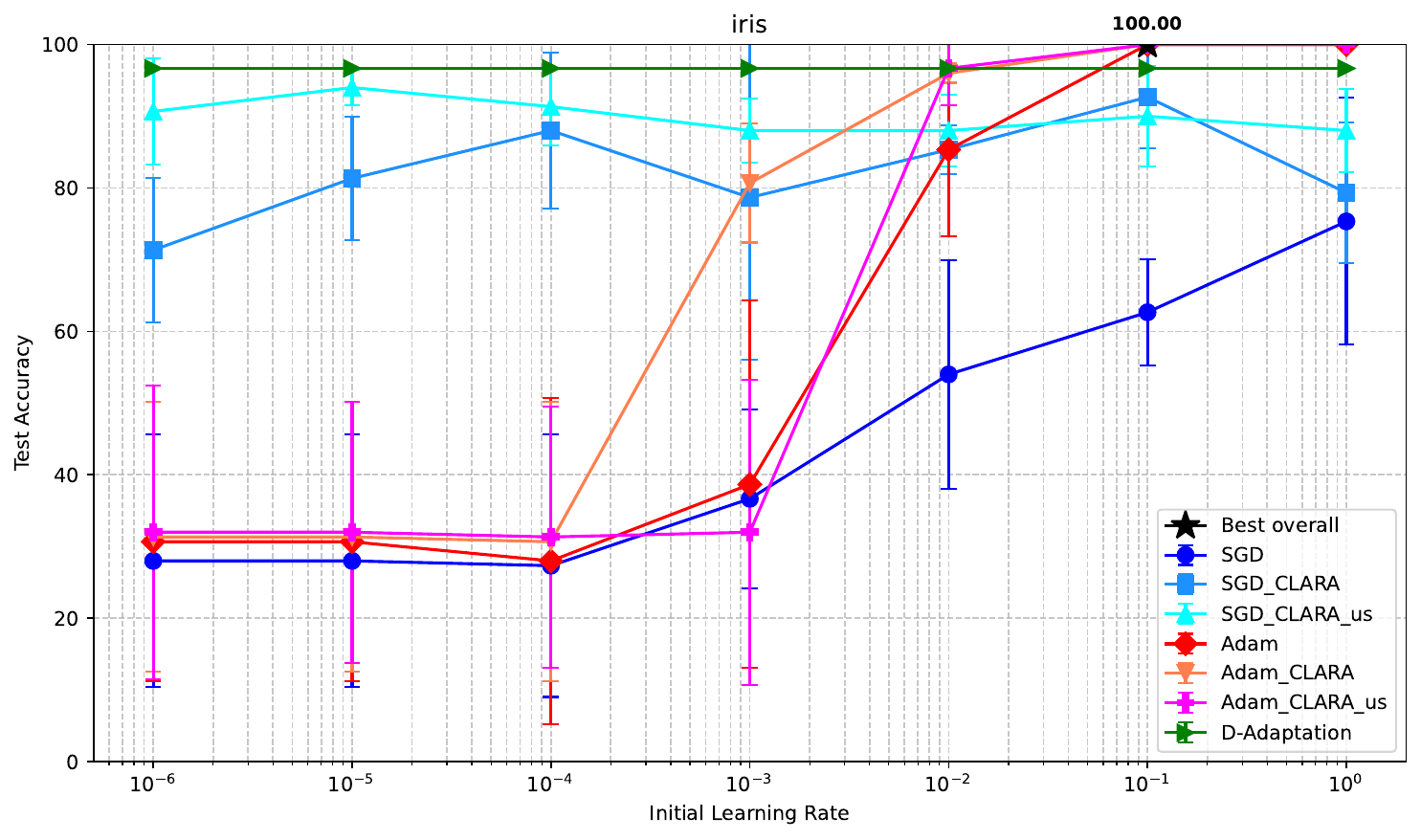}
    \includegraphics[width=0.49\linewidth]{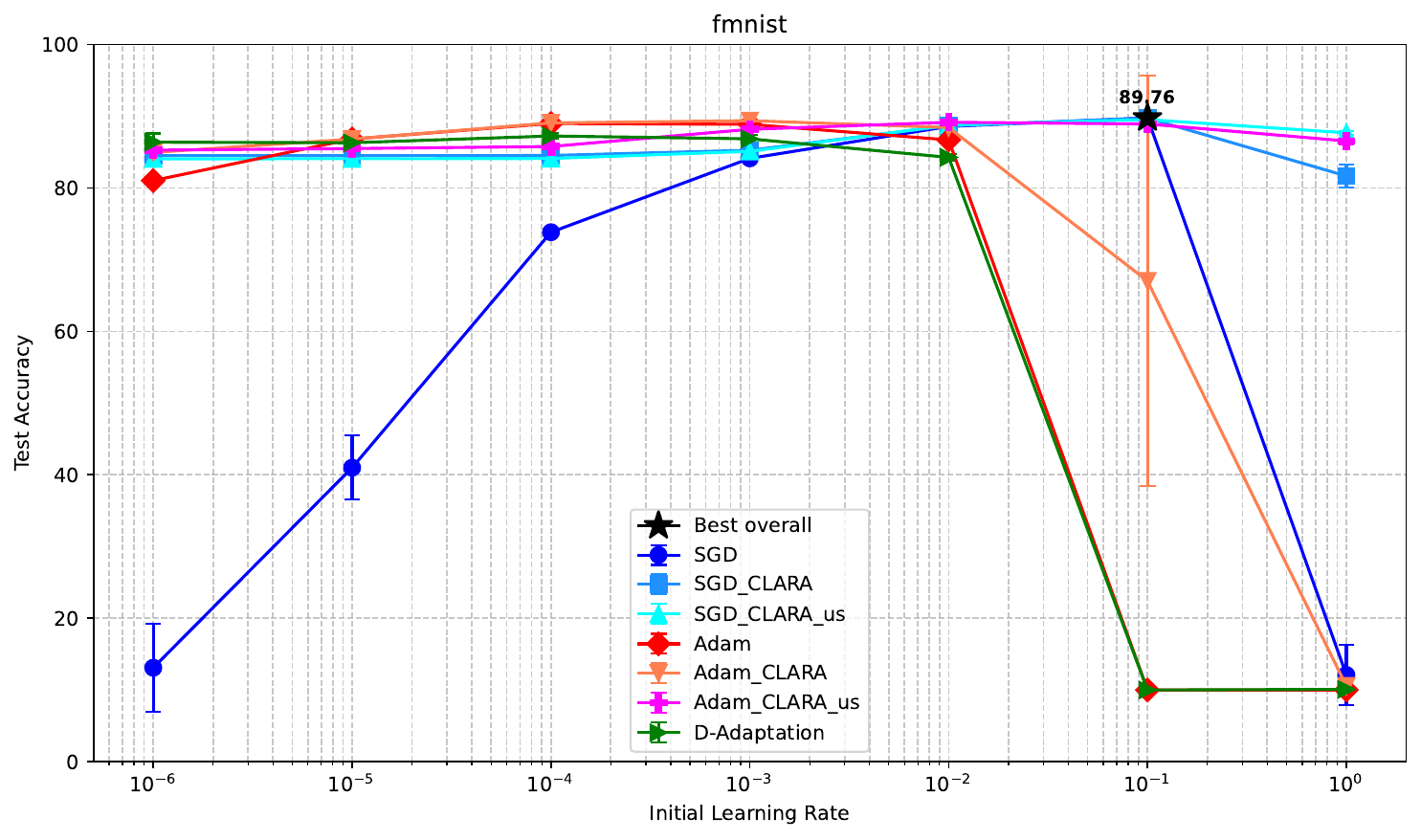} \\
    \includegraphics[width=0.49\linewidth]{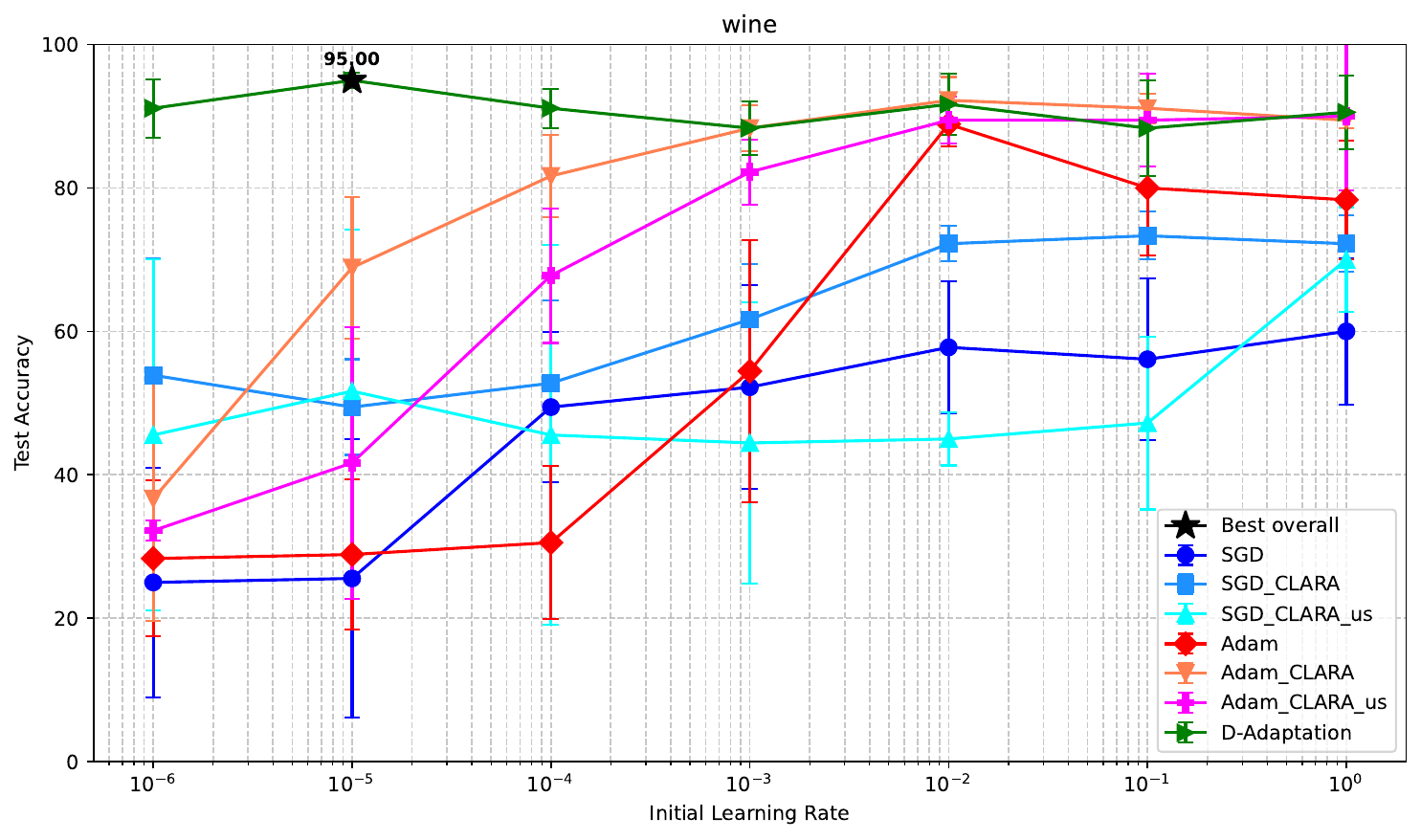}
    \includegraphics[width=0.49\linewidth]{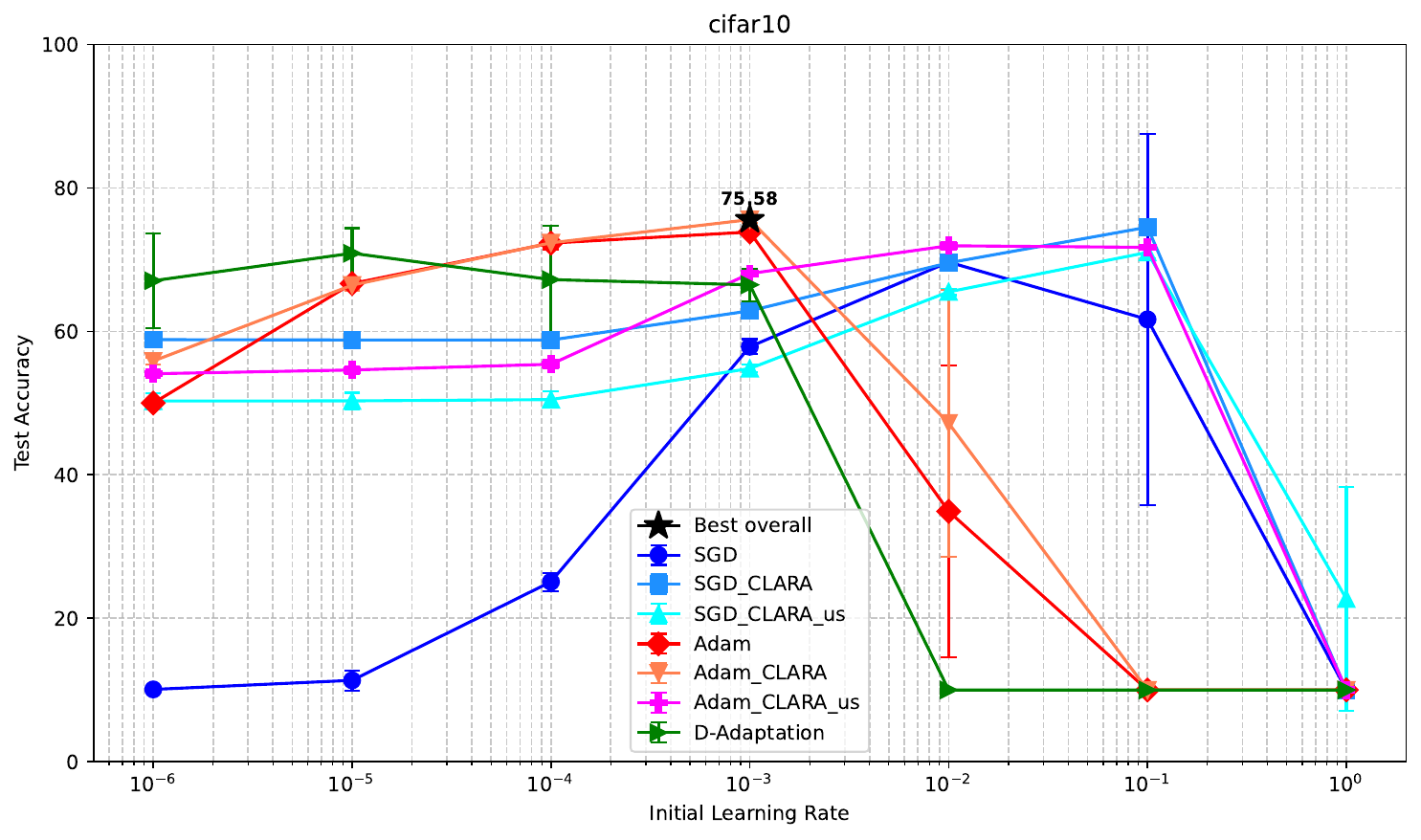} \\
    \includegraphics[width=0.49\linewidth]{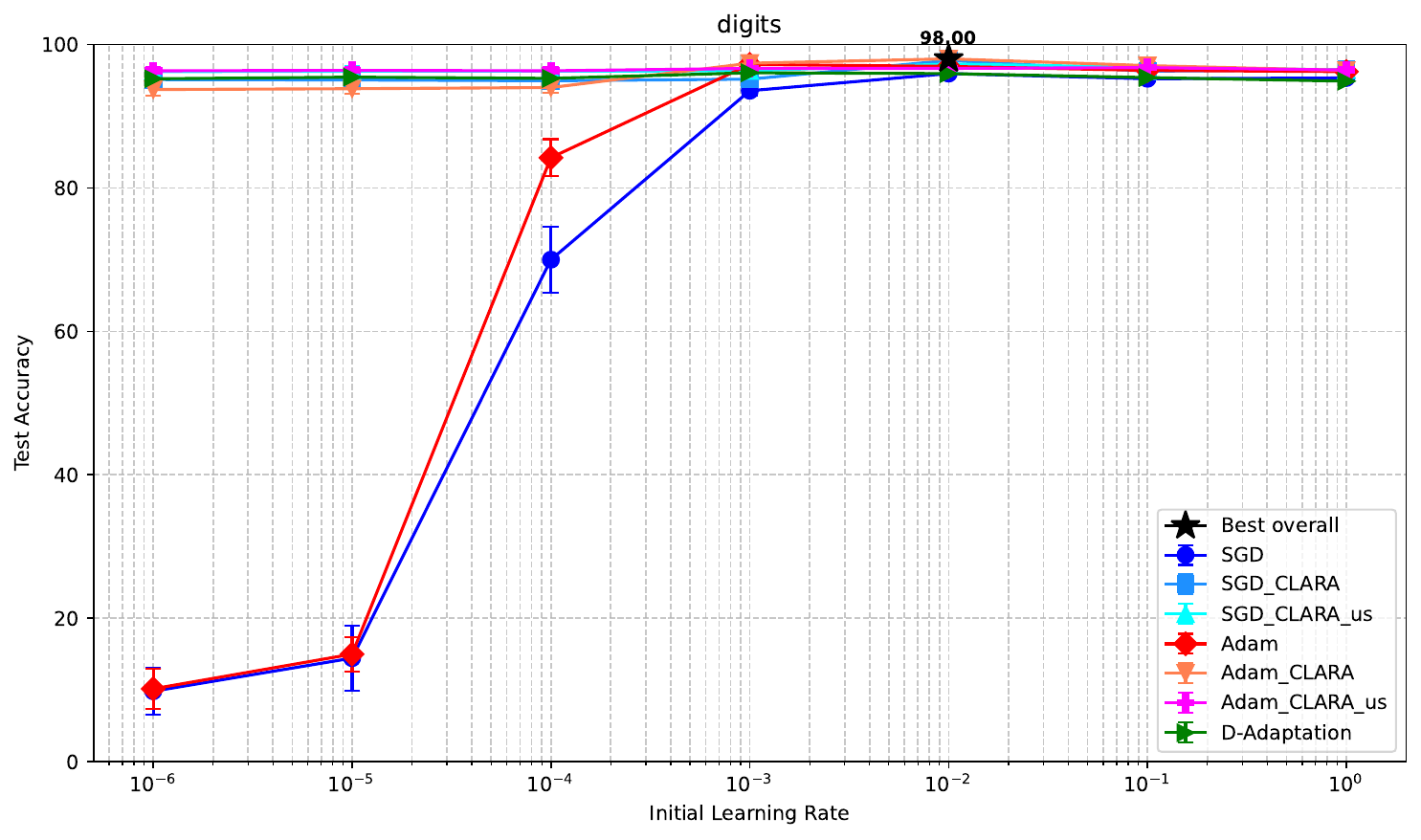}
    \includegraphics[width=0.49\linewidth]{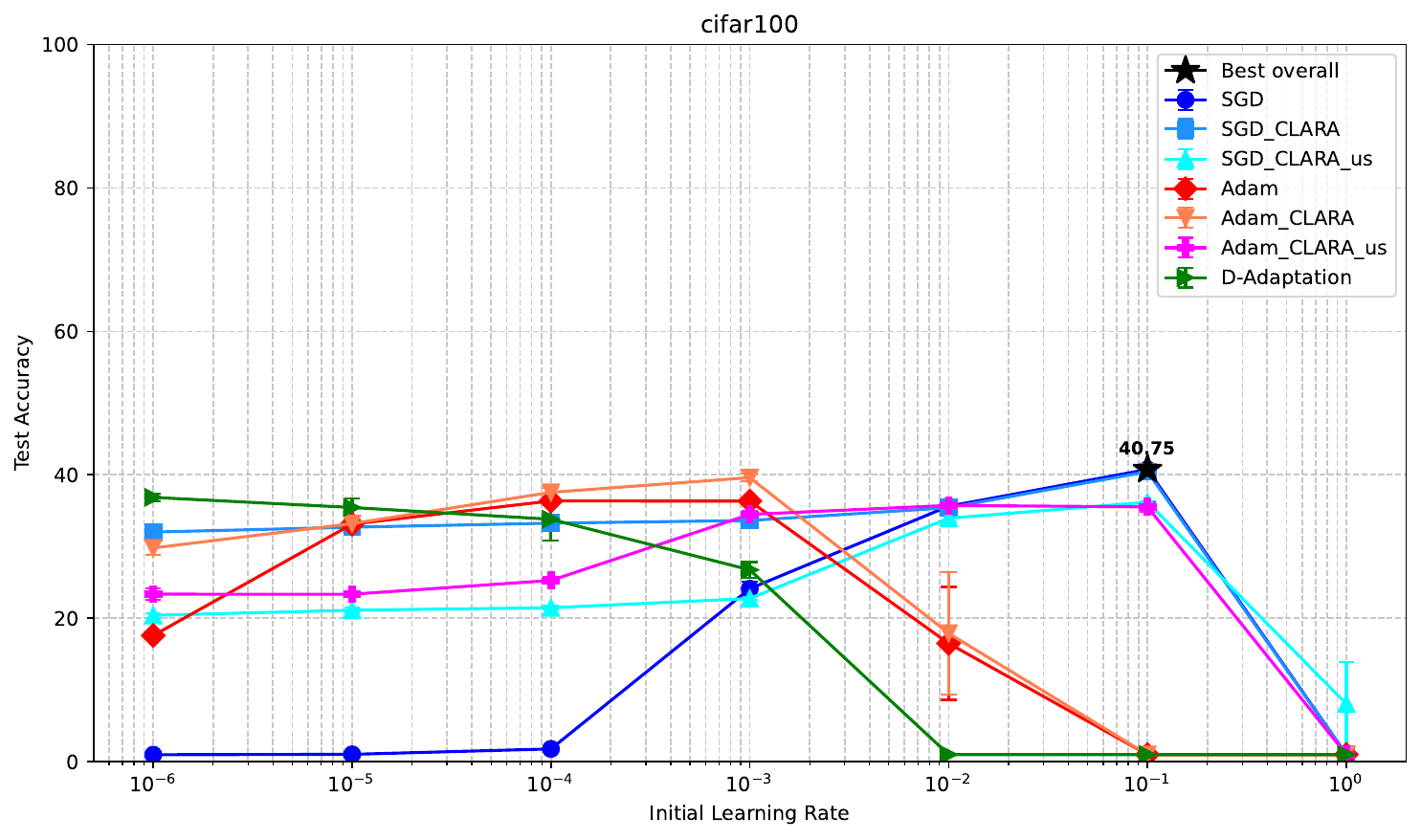}
    \caption{Test accuracy versus initial learning rate for all algorithms, averaged over five seeds. Each plot corresponds to a different dataset. Error bars represent the standard deviation across seeds. For CLARA variants, the best-performing damping value is selected per configuration (dataset and learning rate). While baseline optimizers are sensitive to the choice of initial learning rate, CLARA-enhanced variants show improved robustness across multiple scenarios. The best overall result per dataset is marked with a star.}
    \label{fig:test-accuracy-all-algorithms}
\end{figure}

\paragraph{Comparative performance across learning rates}
The test accuracy was averaged over five seeds as a function of the initial learning rate $\eta_0 \in \{10^{-6}, \dots, 1\}$ for all algorithms and data sets, which is shown in Figure~\ref{fig:test-accuracy-all-algorithms}. CLARA variants demonstrate greater robustness to the choice of $\eta_0$ compared to their static counterparts. In particular, SGD\_CLARA and SGD\_CLARA\_us consistently outperform SGD when the initial learning rate is suboptimal (either too small or too large), recovering performance in regions where vanilla SGD fails to converge.
The picture is more nuanced for Adam. Although the improvements are generally more modest than for SGD, there is consistently at least one Adam\_CLARA variant that outperforms baseline Adam. The performance gains tend to be moderate, and robustness to learning rate initialization is less consistent across settings. This may be due to the internal moment-based rescaling Adam already performs, which partially compensates for poor learning rate initialization. However, cases such as the noisy ellipsoid demonstrate that even Adam’s preconditioning is not always sufficient to prevent divergence, while CLARA-guided adaptation can restore convergence.
D-Adaptation performs particularly well on the convex benchmarks, which aligns with its design: it approximates the optimal learning rate for convex Lipschitz objectives. However, its performance (and that of other adaptive methods) tends to degrade when the initial learning rate is too large. In such cases, the optimizer may quickly enter unstable regions of the loss landscape, from which recovery is difficult even with dynamic learning rate adjustment. This effect is especially visible on the high-curvature, non-convex image benchmarks.

\begin{figure}[t]
    \centering
    \begin{subfigure}[b]{0.99\linewidth}
        \includegraphics[width=0.49\linewidth]{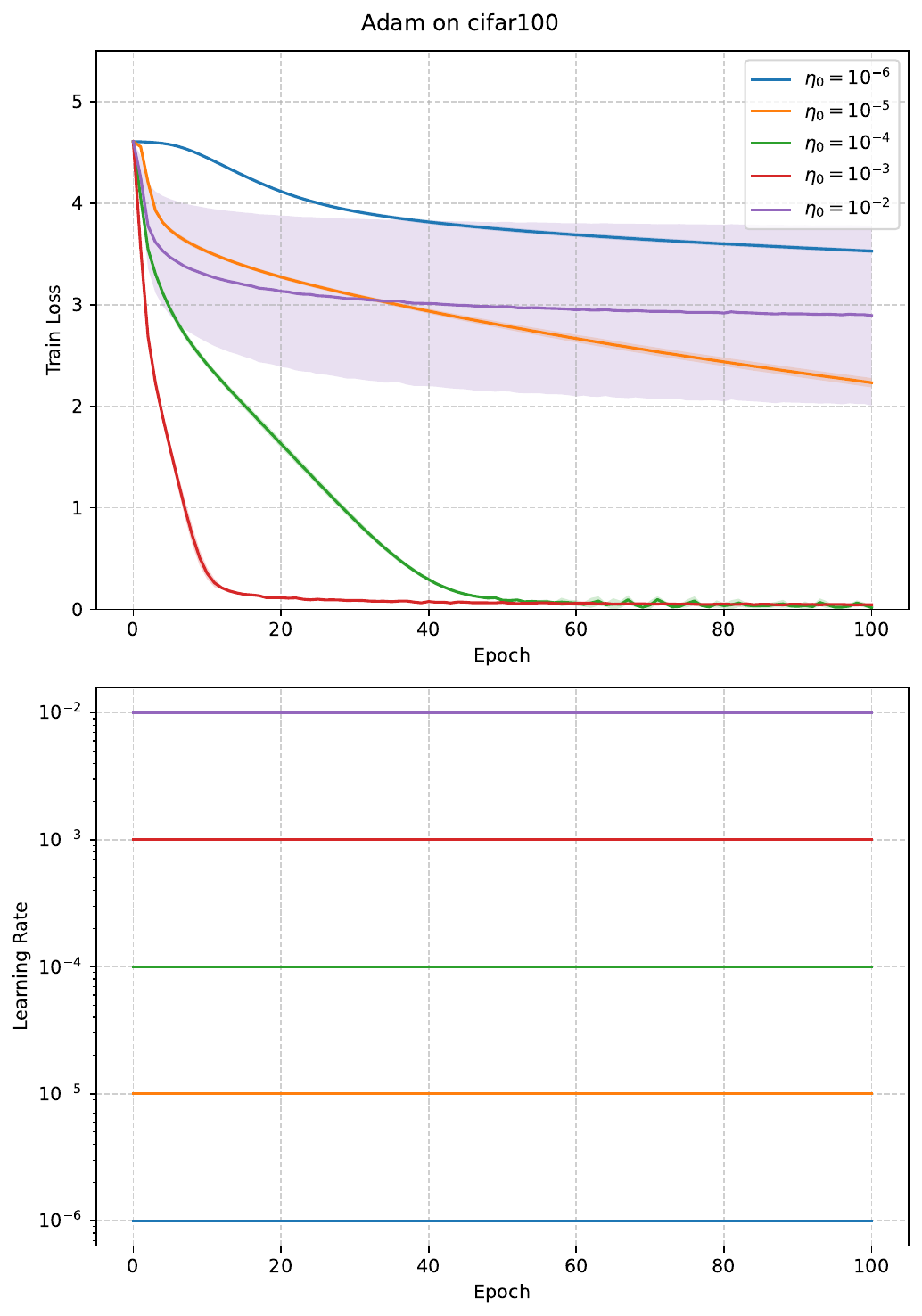}
        \includegraphics[width=0.49\linewidth]{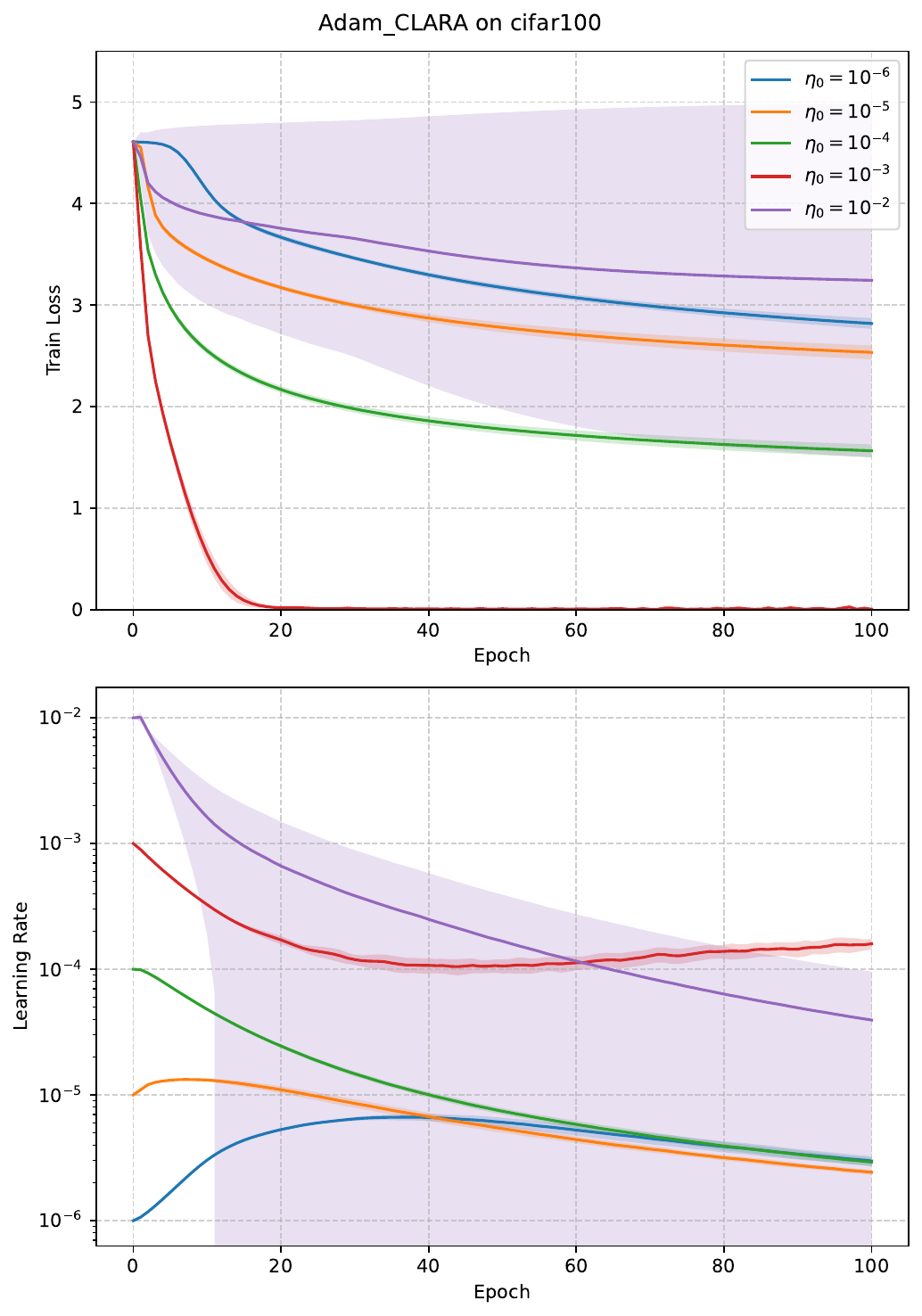}
        \caption{Adam and Adam\_CLARA}
        \label{fig:adam-vs-clara-cifar100}
    \end{subfigure}
    \hfill
    \begin{subfigure}[b]{0.99\linewidth}
        \includegraphics[width=0.49\linewidth]{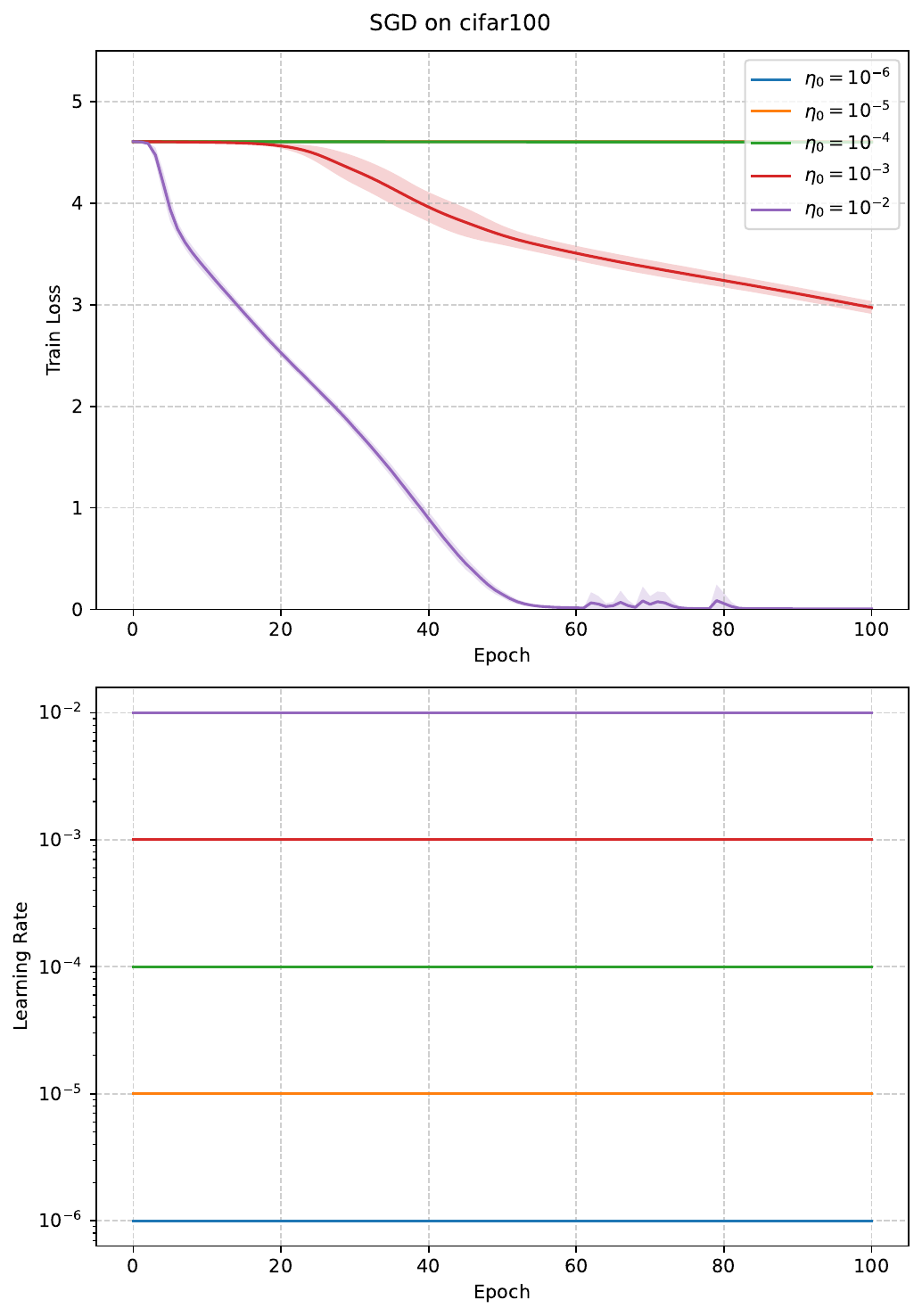}
        \includegraphics[width=0.49\linewidth]{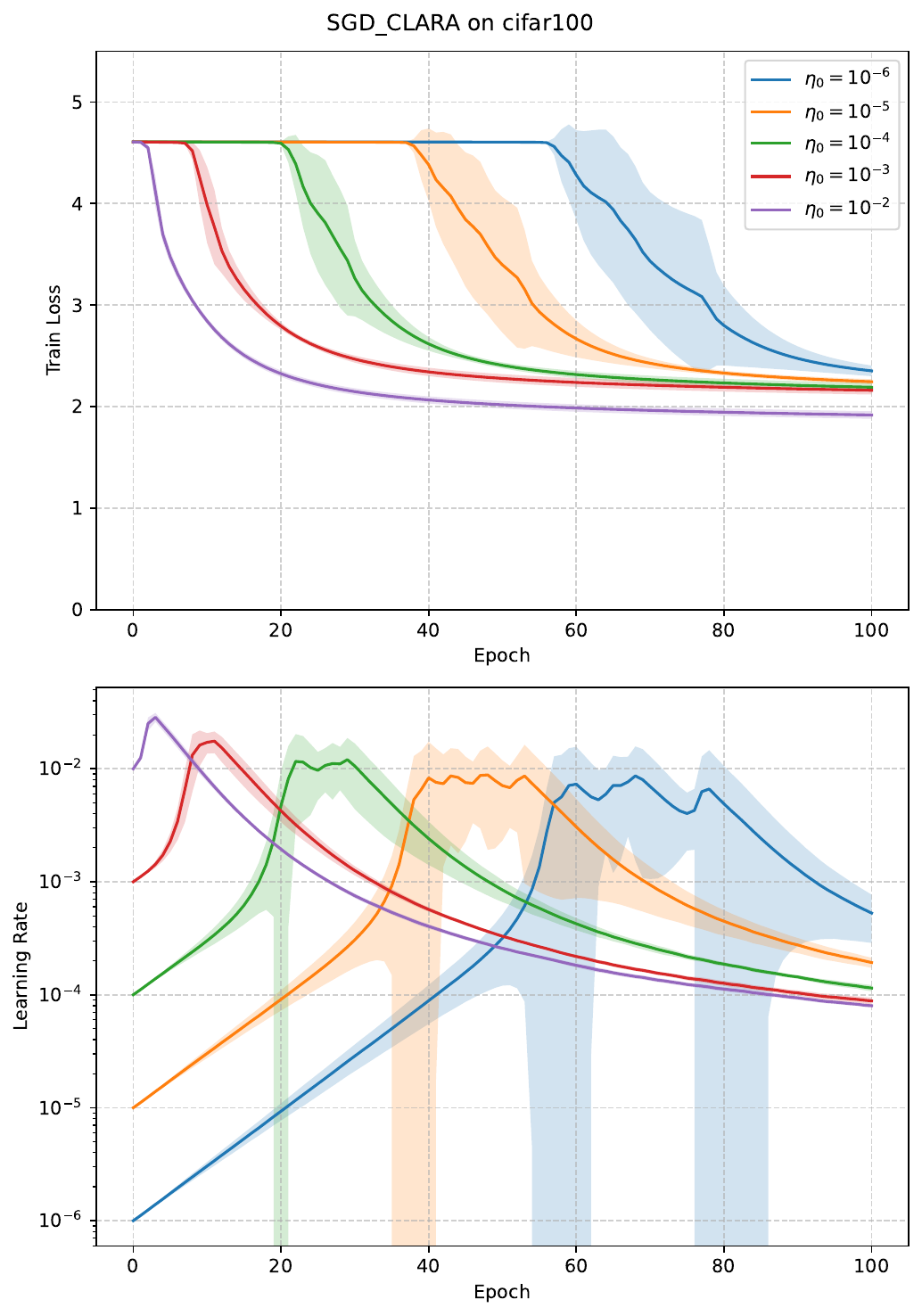}
        \caption{SGD and SGD\_CLARA}
        \label{fig:sgd-vs-clara-cifar100}
    \end{subfigure}
    \caption{Training loss (top) and learning rate (bottom) averaged over five seeds for Adam and SGD with and without CLARA on CIFAR-100, across various initial learning rate values. The damping value used for CLARA is $d = 10^{-3}$.}
    \label{fig:learning-curves-cifar100}
\end{figure}

\paragraph{Learning rate dynamics with CLARA}
To better understand how CLARA adapts learning rates during training, we examine learning curves on CIFAR-100 (Figure~\ref{fig:learning-curves-cifar100}), showing both training loss and the evolution of the effective learning rate over time. For each algorithm, we track performance for several $\eta_0$ values, using a fixed damping value $d = 10^{-3}$.
For SGD\_CLARA, the adaptation patterns are consistent and interpretable: when $\eta_0$ is too small, the learning rate increases early on, eventually stabilizing around a more effective value. This typically coincides with the steepest loss descent. Later, as the optimizer approaches a solution, the learning rate gradually decreases, effectively entering a ``refinement'' phase. This behavior closely resembles a hand-tuned learning rate schedule but emerges automatically.
For Adam\_CLARA, learning rate dynamics are more dataset- and initialization-dependent. While in some cases adaptation helps recover performance ($\eta_0 = 10^{-6}$), and accelerates convergence for favorable initializations ($\eta_0 = 10^{-3}$), the learning rate trajectories vary more and offer less clear interpretability than with SGD\_CLARA. Nonetheless, these curves match the best-performing configurations of Adam in most settings.
Overall, these observations reinforce the finding that CLARA offers greater value when applied to SGD, where no intrinsic adaptation is available. For Adam, CLARA can still offer recovery from overly conservative initializations, but its benefits are more context-dependent. 

%-----------------------------------%
\section{Discussion and Conclusion}
\label{sec:discussion}
%-----------------------------------%

The choice and adjustment of the learning rate remains one of the most critical yet under-specified aspects of training deep learning models. Despite the popularity of adaptive optimizers like Adam and the widespread use of heuristic schedules, a robust, general-purpose solution for online learning rate control is still lacking. This work explores an alternative through CLARA, a cumulative adaptation mechanism that dynamically modulates the global learning rate based on the optimizer's trajectory.

CLARA builds on a path length–based adaptation scheme originally proposed for SGD~\cite{SchoenauerSebag2017}, but extends and corrects it for use with optimizers like Adam which internally rescale gradient directions. In particular, we address the conceptual mismatch that arises when applying the original method to optimizers that alter gradient directions, and we introduce a formulation that reflects Adam’s internal geometry more faithfully. We evaluate both SGD- and Adam-based CLARA variants across synthetic landscapes and standard supervised learning tasks.

Empirically, CLARA proves most beneficial when applied to SGD, offering consistent robustness to poor initializations and often recovering from suboptimal learning rates without manual tuning. Its behavior closely resembles that of hand-designed schedules, but emerges automatically from alignment statistics between steps. For Adam, the benefits are more context-dependent, as the optimizer already adjusts step magnitudes internally. Still, a properly corrected CLARA variant often matches or improves performance, particularly in regimes where fixed learning rates fail.
We also benchmarked against D-Adaptation~\cite{defazio2023}, which performs well on convex tasks, reflecting its theoretical foundation, but is more fragile under large initial learning rates and in non-convex settings. In contrast, CLARA exhibits more stable behavior across problem types and initialization ranges.

Overall, CLARA offers a lightweight, general-purpose mechanism for learning rate control that improves optimizer robustness with minimal overhead. While not a universal replacement for tuning, it provides a practical safeguard against poorly chosen hyperparameters and lays the groundwork for more principled, trajectory-aware adaptation strategies.
Future work could explore automated strategies for setting the damping parameter, potentially making CLARA fully parameter-free. Evaluating CLARA in reinforcement learning or continual learning settings, where stability and adaptivity are especially critical, also remains a promising direction for future research.

\bibliographystyle{plain}
\bibliography{tm_bib}

\end{document}